\documentclass{ecai} 
\usepackage{caption}
\newcommand{\githuburl}{https://github.com/anonymized/chbs}

\usepackage{latexsym}
\usepackage{amssymb}
\usepackage{amsmath}
\usepackage{amsthm}
\usepackage{booktabs}
\usepackage{enumitem}
\usepackage{graphicx}
\usepackage{color}

\usepackage{hyperref}
\usepackage{multirow}

\usepackage{array}
\usepackage{textcomp}
\usepackage{url}
\usepackage{verbatim}

\usepackage{makecell}

\usepackage{cleveref}
\usepackage{color}
\usepackage{epstopdf}
\usepackage{algpseudocode}
\usepackage{comment}
\usepackage{tikz}
\usepackage{makecell}
\usepackage{nomencl}
\usepackage[utf8]{inputenc}

\usepackage{placeins}

\usepackage{graphicx}

\usepackage{subcaption}

\newcommand{\BibTeX}{B\kern-.05em{\sc i\kern-.025em b}\kern-.08em\TeX}

\begin{document}


\begin{frontmatter}


\paperid{123} 


\title{Supervised Embedded Methods \\for Hyperspectral Band Selection}


\author[A]{\fnms{Yaniv}~\snm{Zimmer}\orcid{0009-0007-4342-8866}}
\author[A]{\fnms{Ofir}~\snm{Lindenbaum}\orcid{0000-0002-5990-742X}}
\author[A]{\fnms{Oren}~\snm{Glickman}\orcid{0009-0000-5158-7372}\thanks{Corresponding Author. Email: oren.glickman@biu.ac.il}} 

\address[A]{Bar-Ilan University}


\begin{abstract}
Hyperspectral Imaging (HSI) captures rich spectral information across contiguous wavelength bands, supporting applications in precision agriculture, environmental monitoring, and autonomous driving. However, its high dimensionality poses computational challenges, particularly in real-time or resource-constrained settings. While prior band selection methods attempt to reduce complexity, they often rely on separate preprocessing steps and lack alignment with downstream tasks.
We propose two novel supervised, embedded methods for task-specific HSI band selection that integrate directly into deep learning models. By embedding band selection within the training pipeline, our methods eliminate the need for separate preprocessing and ensure alignment with the target task.
Extensive experiments on three remote sensing benchmarks and an autonomous driving dataset show that our methods achieve state-of-the-art performance while selecting only a minimal number of bands.
These results highlight the potential of efficient, task-specific HSI pipelines for practical deployment.
\end{abstract}

\end{frontmatter}


\section{Introduction}
Hyperspectral imaging (HSI) captures reflectance across hundreds of contiguous spectral bands, enabling fine-grained material differentiation beyond conventional RGB imaging. This capability has driven advances in multiple application areas. In autonomous driving, HSI helps distinguish between road markings, surface types, and vegetation, enhancing semantic segmentation and safety in Advanced Driver-Assistance Systems (ADAS) and Autonomous Driving Systems (ADS) \cite{huang2021weakly-4chip, colomb2019spectral-2chip, weikl2022potentials-3chip}. In agriculture, for example, it enables early disease detection and crop classification, supporting precision farming \cite{Fass2024}.

However, the high dimensionality of HSI introduces significant challenges for computation, storage, and transmission, especially in real-time or resource-limited scenarios. Band selection addresses this by identifying informative spectral bands that preserve task-relevant information, and can also inform the design of simplified, application-specific sensors \cite{pinchon2019all-chip5,winkens2019deep-chip6}. However, existing methods often rely on separate preprocessing and prioritize spectral reconstruction over task utility.

To address these limitations, we propose two embedded supervised band selection methods (Embedded Hyperspectral Band Selection, EHBS, and Concrete Hyperspectral Band Selection, CHBS) based on reparameterization techniques. These integrate band selection directly into the training pipeline, eliminating preprocessing and optimizing for downstream tasks. A dedicated selection layer positioned between the model’s input and the first layer of the downstream task deep learning model, enables adaptive and efficient feature selection.
We evaluate our methods on three remote sensing benchmarks and one autonomous driving dataset. Our approaches consistently outperform existing methods, especially in low-band scenarios, with CHBS excelling in resource-constrained settings by dynamically adapting to task needs.

These results establish a scalable, task-adaptive framework for efficient HSI processing, with strong potential for real-world deployment in autonomous systems and environmental sensing.

\section{Related Work}
\label{sec:related_work}
\noindent Hyperspectral band selection methods can be broadly classified by their use of supervision (unsupervised vs. supervised) and whether selection is embedded in model training or performed as a preprocessing step. This section summarizes key approaches and their limitations.
Unsupervised methods use statistical or information-theoretic criteria without task labels. PCA-based techniques \cite{kang2017pca, sun2018graph} reduce dimensionality by preserving variance but lack task specificity. Reconstruction-based approaches (e.g., BS-Nets~\cite{bs_nets}, DARecNet-BS~\cite{roy2020darecnet}, and TAttMSRecNet~\cite{TAttMSRecNet}) employ autoencoders to reconstruct spectra from selected bands, optimizing fidelity over utility. Sparsity-based methods such as SpaBS~\cite{sun2014new} and SNMF~\cite{sun2015band} identify representative bands via sparse modeling or clustering, but typically overlook downstream performance.

Supervised methods use labels to guide selection, improving task relevance. However, many are non-embedded—executed as separate preprocessing—adding complexity. Genetic algorithms~\cite{10078841, esmaeili2023hyperspectral, geneticalgorithm},  concrete autoencoders ~\cite{DBLP:journals/corr/abs-1901-09346, sun2021novel}, and deep reinforcement learning (DRL) approaches~\cite{mou2021deep, feng2021deep, feng2024multi} have shown promise, but often incur high computational cost and focus on spectral reconstruction rather than direct task optimization.
Existing methods share key drawbacks: (1) optimizing spectral fidelity rather than task utility, (2) decoupling selection from training, and (3) inefficiency when selecting few bands for real-time use.
To address these gaps, we propose two embedded, supervised methods—EHBS and CHBS. EHBS employs group-wise gating and \(\ell_0\) regularization, while CHBS uses a Concrete Selector Layer and Gumbel-Softmax reparameterization to enable differentiable band selection. Both approaches integrate directly into model training, enabling task-aware, adaptive feature selection without separate preprocessing.

\section{Problem definition}
\noindent Hyperspectral band selection aims to reduce the dimensionality of hyperspectral data by selecting the most informative spectral bands for a specific downstream task. This section formalizes the task as a joint optimization problem, combining band selection with model learning to enable seamless integration into task-specific models.

Let \(X\) denote a dataset of \(m\) instances, where each instance \(x_i \in \mathbb{R}^{n \times h \times w}\) represents a hyperspectral image with \(n\) spectral bands and spatial dimensions \(h \times w\). Corresponding labels for the downstream task are given by \(Y = \{y_i\}_{i=1}^m\), where \(y_i\) is the label for \(x_i\). Here, \(n\) represents the total number of spectral bands.

Band selection is represented by an indicator vector \(\mathbf{I} \in \{0,1\}^n\), where \(\mathbf{I}_j = 1\) indicates that band \(j\) is selected for processing, and \(\mathbf{I}_j = 0\) otherwise. The $l_0$ norm of an indicator function $\mathbf{I}$ corresponds to the number of selected bands where \(\|\mathbf{I}\|_0 = k\) specifies that exactly \(k\) bands are selected. The selected bands are applied to each instance \(x_i\) via an element-wise product \(x_i \odot \mathbf{I}\), effectively masking unselected bands.

Let \(F\) denote a family of models for the downstream task, parameterized by \(\theta\), and let \(Loss\) represent a task-specific loss function. The objective is to jointly optimize the band selection vector \(\mathbf{I}\) and the model \(f_\theta \in F\) to minimize the average loss over the dataset:
\begin{equation}
\label{eq:optimization}
\arg \min_{\theta, \mathbf{I}, \|\mathbf{I}\|_0 = k} \frac{1}{m} \sum \limits_{i=1}^m Loss(f_{\theta}(x_i \odot \mathbf{I}), y_i).
\end{equation}

This formulation ensures that the selected bands are directly optimized for the downstream task, enabling efficient and task-specific dimensionality reduction. Unlike traditional preprocessing approaches, this embedded strategy integrates band selection into the model training process, improving performance and adaptability.

\section{Proposed Methods}
\label{sec:methods}
\noindent We propose two novel, embedded feature selection methods for task-specific hyperspectral band selection, designed to integrate seamlessly with deep learning models. Both methods leverage the reparameterization trick  \cite{jang2016categorical, kingma2013auto, kingma2015variational} to address the challenge of \(\ell_0\) regularization, which is non-differentiable and thus difficult to optimize in standard deep learning frameworks. By incorporating a dedicated selection layer that dynamically determines the importance of each spectral band during training, our approaches eliminate the need for separate preprocessing and enable end-to-end optimization of both band selection and downstream model parameters.

\subsection{EHBS Method: \texorpdfstring{\(\ell_0\)}{L0}-Based Regularization}
\label{sec:EHBS}
\noindent The EHBS method, based on the Stochastic Gates framework \cite{pmlr-v119-yamada20a}, leverages the reparameterization trick to
enable differentiable learning of band importance probabilities. 
These probabilities are then relaxed into binary values during
training using an \(\ell_0\) approximation, effectively performing
band selection as part of the model’s training process.

\begin{figure}[!htb]
    \begin{center}
    \includegraphics[width=0.99\linewidth]{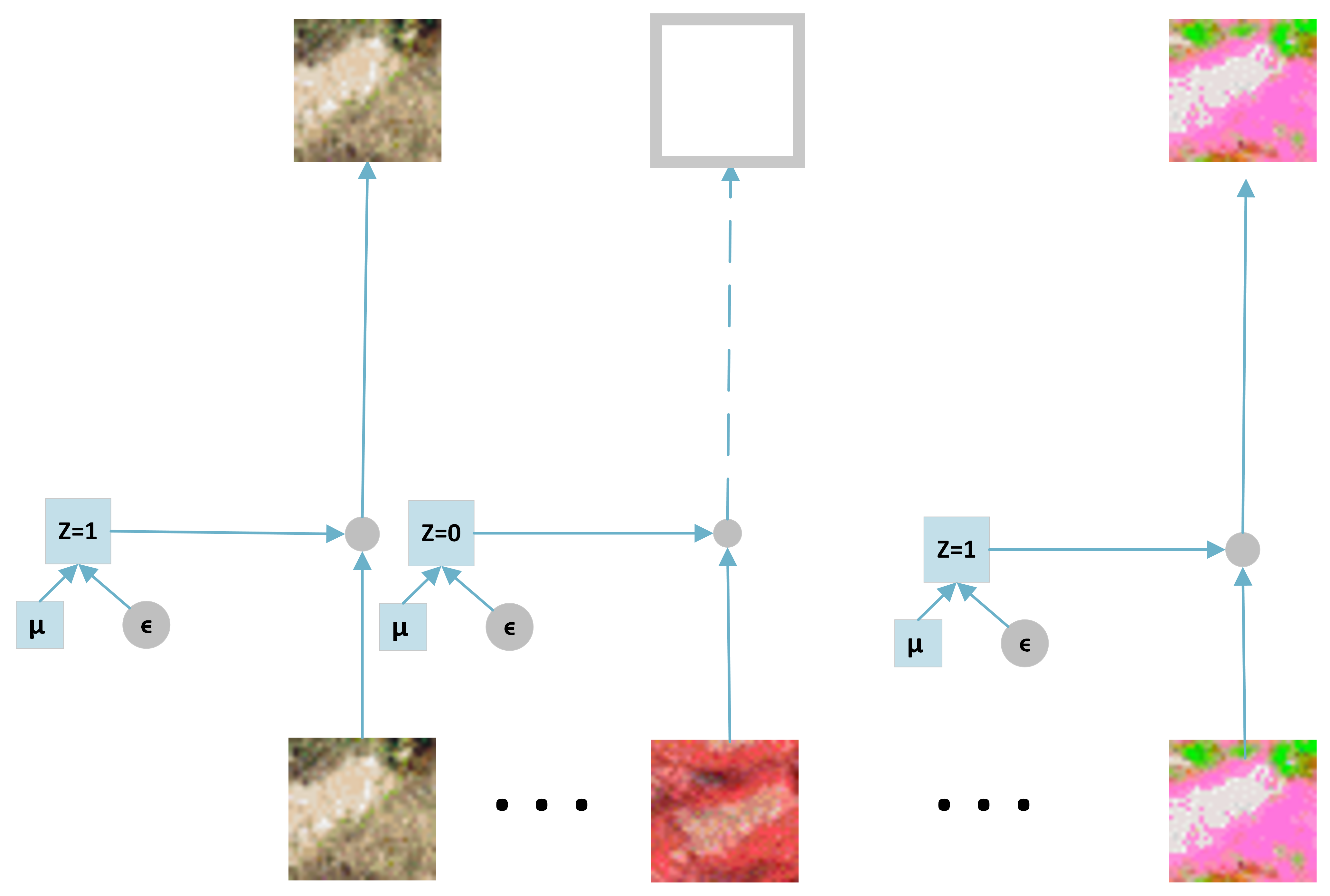}
    \caption{Illustration of the EHBS layer, showing the gating mechanism for selecting spectral bands based on task-specific importance.}
    \label{fig:EHBS_layer}
    \end{center}
\end{figure}

The EHBS layer consists of \(n\) stochastic gates, where \(n\) corresponds to the total number of spectral bands (see Figure~\ref{fig:EHBS_layer}). Each gate includes a learnable parameter \(\mu_i\) that indicates the importance of its associated band. 
During the forward pass, each gate computes its value \(z_i\) as:
$z_i = \text{clamp}(\mu_i + \epsilon_i)$,
where \(\epsilon_i \sim \mathcal{N}(0, \sigma^2)\) is Gaussian noise, and \(\text{clamp}(x) = \max(0, \min(x, 1))\) ensures the value lies in \([0, 1]\). This value \(z_i\) is used to mask the corresponding input band, thereby retaining all features of the band or suppressing them entirely. The noise facilitates exploration during training and is fixed during inference.

To drive the gate values toward a binary decision, a regularization term, $R$, approximating the \(\ell_0\) norm is added to the loss function:
\[
R = \lambda \sum_{i=1}^{n} \Phi\left(\frac{\mu_i}{\sigma}\right),
\] 
where \(\Phi\) denotes the standard Gaussian cumulative distribution function, \(\mu_i\) is the mean of gate \(i\), and \(\lambda\) is the regularization coefficient. This term approximates the \(\ell_0\) norm, guiding the gates toward sparse band selection.

To meet a target number of selected bands \(k\), we adapt the stochastic gates framework by modifying how non-selected features are handled. Unlike the traditional stochastic gates framework, where unselected features are assigned zero weight, we implement a selector that explicitly chooses the top \(k\) bands. This adaptation is crucial since our downstream task includes convolutional layers, which can be negatively affected by zeroed bands changing dynamically during training. First, the regularization coefficient  \(\lambda\) is set based on the target number of bands \(k\); a higher \(\lambda\) is used when \(k\) is small to enforce stronger sparsity. The selection layer then ranks all bands based on their \(\mu\) values and selects the top \(k\) bands. To maintain spectral coherence, the selected bands are ordered by their spectral values before being passed as output. This ordering leverages domain knowledge of the contiguous nature of spectral bands, ensuring a consistent input structure that enhances model stability and interpretability.

\subsection{CHBS Method: Concrete Selector Layer}
\label{sec:concrete}
\noindent The CHBS method applies the reparameterization trick by incorporating a Concrete Selector Layer for band selection, adapted from the feature selection approach proposed in \cite{jang2017categorical}. This technique enables the approximation of discrete band selection in a continuous and differentiable manner, facilitating seamless integration with deep learning models.

The layer is defined by the parameters \((L, \tau, \alpha, \beta)\):  
\begin{itemize}
    \item \(L \in \mathbb{R}^{k \times n}\): is a learnable logits matrix where \(n\) is the total number of bands and \(k\) is the target number of selected bands. Each $n$-dimensional row of $L$ corresponds to a selector for a particular band.
    \item \(\tau \in (0, \infty)\): is a temperature parameter that controls the smoothness of the Gumbel-Softmax distribution, initialized with a fixed value and decayed during training by a factor \(\alpha\).
    \item \(\alpha\): The decay factor for \(\tau\), enabling a gradual transition from a soft to a nearly discrete (one-hot) representation.
    \item \(\beta\): is the scale parameter of the Gumbel distribution, controlling the magnitude of the noise added to the logits when calculating the mask vectors in the forward pass.
\end{itemize}

\noindent At each forward pass, the calculated Gumbel softmax per each row of the matrix is multiplied by the full HSI 3D input and is passed on to the first layer of the deep learning network of the downstream task. Given an input tensor \(\mathbf{x} \in \mathbb{R}^{n \times h \times w}\), where \(n\) is the number of bands and each band is an image of size \((h, w)\), 
the encoder encodes \(\mathbf{x}\) into a \(\mathbb{R}^{k \times h \times w}\) as a weighted sum of input bands by multiplying it by a selection matrix \(\mathbf{M} \in \mathbb{R}^{k \times n}\).
The matrix $M$ is derived from the logit matrix $L$ and the temperature parameter $\tau$ by applying the Gumbel-Softmax operation over the rows of $L$ as follows:
\[
M_{i,j} = \frac{\exp\left((L_{i,j} + G_{i,j})/\tau\right)}{\sum_{r=1}^{n} \exp\left((L_{i,r} + G_{i,r})/\tau\right)},
\]
where $G_{i,j} = -\log(-\log(u_{i,j}))$ and $u_{i,j} \sim \text{Uniform}(0,\beta)$ and $\beta$ is the scale parameter of the Gumbel distribution that controls the magnitude of the added noise.

The parameters \((L, \tau, \alpha, \beta)\) must be initialized before training, as they significantly influence model performance and convergence. The values of the matrix $L$ are learned as part of the training process of the entire neural network in an attempt to solve the optimization problem defined in~\eqref{eq:optimization}.
As $\tau \to 0$ ( by multiplying it by the decay factor $\alpha$ at the end of each batch), the distribution smoothly approaches a discrete one-hot vector
This reparameterization allows for differentiable sampling, enabling gradient-based optimization during the training process.
During inference, the noise is not applied; instead, a corresponding one-hot encoded matrix derived from the logits matrix is used to make a final selection efficiently and deterministically.  

\subsubsection{Initialization of the logits matrix}
\label{sec:logits_initialization}
\noindent We propose Segmented Xavier Initialization, a novel scheme specifically designed for hyperspectral band selection. Unlike the standard Xavier (Glorot) initialization \cite{pmlr-v9-glorot10a}, Segmented Xavier Initialization initializes the logits matrix \(L\) to target distinct regions of the hyperspectral spectrum. This approach allows the selectors to focus more effectively on relevant spectral bands. Given a \(k \times n\) logits matrix \(L\), where \(n\) represents the total number of spectral bands and \(k\) is the target number of bands, we partition the \(n\) bands spectrum into \(k\) contiguous segments, each of size \(\left\lfloor \frac{n}{k} \right\rfloor\). Each row (gate) in \(L\) is then initialized to emphasize a different segment. This is achieved by modifying the Xavier initialization so that the mean value within the designated focus area is positive, while values outside this area are negative. This adjustment ensures that the overall mean remains at zero and preserves the variance properties of standard Xavier initialization. By structuring the initialization in this way, the Segmented Xavier Initialization provides a guided starting point for each selector, reducing redundancy and improving the stability of band selection.

\begin{figure}[htb]
    \begin{center}
    \includegraphics[width=0.99\linewidth]{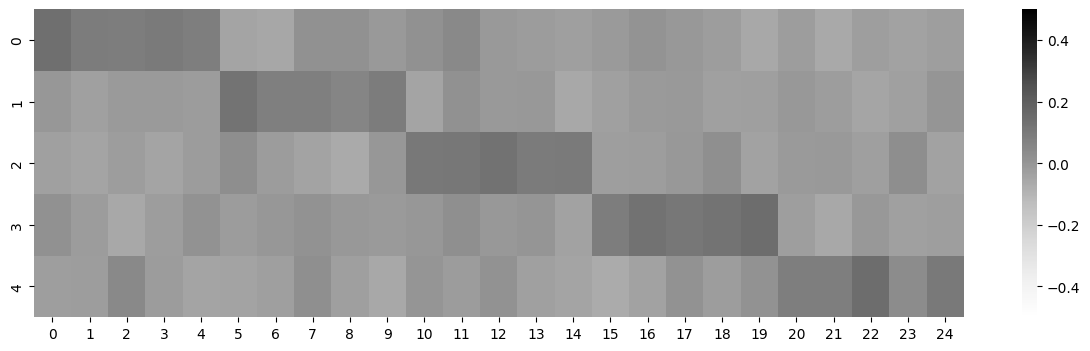}
    \caption{Heatmap visualization of the logits matrix \(L\) after applying the proposed Segmented Xavier Initialization scheme. The figure illustrates the segmentation of 25 bands ($x$-axis) into five target bands, where each selector ($y$-axis) is biased towards a specific spectral region. Each cell corresponds to the initial value of $\mu$, with the heatmap visualization reflecting the distribution of these values. This structured initialization mitigates the risk of selecting redundant bands by providing a well-distributed starting point for each gate.}
    \label{fig:example_initiazation}
    \end{center}
\end{figure}

An example of this initialization approach, when selecting 5 bands from a total of 25, is illustrated in Figure~\ref{fig:example_initiazation}.

\section{Datasets and Experimental Setup}
\noindent This section details the datasets, baseline models, experimental settings, and evaluation metrics used to validate the proposed methods.

\subsection{Datasets}
\label{sec:datasets_and_tasks}
\noindent We evaluated our proposed model on four semantic segmentation datasets: three remote sensing datasets, PaviaU, Salinas, and Chikusei \cite{PaviaU, Salinas, yokoya2016airborne} and HSI-Drive v2, an Autonomous Driving Systems (ADS) dataset \cite{GUTIERREZZABALLA2023102878}. Table~\ref{tab:datasets_table} summarizes the properties of these datasets.

\begin{table*}[ht]
\caption{Datasets Summary}
\label{tab:datasets_table}
\begin{center}
\resizebox{\textwidth}{!}{
    \begin{tabular}{clrcrrl}
    Type & Dataset & Bands&  Spectrum (nm) & Classes &  Samples & Notes \\
    \hline 
    \multirow{3}{*}{\makecell{Remote\\Sensing}} &  PaviaU & 103&  430-860 & 9& 42,776& \multirow{3}{*}{Each sample is a pixel}\\
    & Salinas&  204&  430-2500& 16 &54,129& \\
    & Chikusei&  128&  363-1018& 19 &5,877,195& \\
    \makecell{ADS}  &  HSI-Drive v2 & ~25& 598-976& 10 &756& Each sample is an image\\
    \end{tabular}
}
\end{center}
\end{table*}

The PaviaU dataset captures an urban scene in northern Italy, providing 103 spectral bands in the 430-860 $nm$ range, with approximately 42,000 valid pixels classified into nine categories. The Salinas dataset, focused on agricultural scenes in California, contains 204 bands (after removing water absorption bands) in the 430-2500 $nm$ range, with approximately 54,000 valid pixels and 16 ground-truth classes, and is known for its high spatial resolution (3.7-meter pixels). The Chikusei dataset is a large-scale benchmark covering urban and agricultural areas in Japan, with 128 spectral bands in the range of 363-1018 $nm$ and annotations for 19 classes. 

The HSI-Drive v2 dataset contains 756 images across 25 spectral bands, labeled into 10 classes, and collected under diverse weather conditions. This dataset is intended to evaluate scenarios relevant to autonomous driving, such as low lighting and rainy weather, providing a robust benchmark for real-world ADS tasks. We used the simplified 5-class grouping task, focusing on five key classes (road, road marks, sky, vegetation, and other) for enhanced 
separability, as this was the setting for which classification results were reported in the dataset paper.

\subsection{Baseline Band Selection Methods}
\noindent To compare our proposed methods with state-of-the-art approaches, we implemented nine different band selection methods that have achieved top performance in recent literature
and that cover the different family types described in \cref{sec:related_work}. The list of methods used for comparison is summarized in Table ~\ref{tab:baseline_methods}.   

\begin{table}[ht]
    \centering
    \resizebox{\linewidth}{!}{%
    \begin{tabular}{lllc}
Method & Family	& Ref.\\ \hline	\hline		
CHBS & Embedded encoder & (Ours)\\
EHBS    &Regularization Deep Learning &  (Ours)\\ 
SNMF & Sparsity and Clustering & \cite{sun2015band}\\
Genetic	& Supervised genetic optimization & \cite{geneticalgorithm} \\
BS-Net-Conv	& Autoencoder Reconstruction	& \cite{bs_nets} \\ \hline
TAttMSRecNet & Autoencoder Reconstruction &	\cite{TAttMSRecNet}	\\ 
DARecNet-BS	&Autoencoder Reconstruction	& \cite{roy2020darecnet} \\
DRL	 & Reinforcement Deep Learning & \cite{mou2021deep}\\
PCA		& Dimension Reduction & \cite{sun2018graph}	\\
SpaBS	& Sparsity & \cite{sun2014new}

\end{tabular}
}
    \caption{Baseline band selection methods}
    \label{tab:baseline_methods}
\end{table}

All methods were applied to the PaviaU and Salinas remote sensing datasets. For the larger datasets (Chikusei and HSI-Drive v2), we limited comparison to the top three methods that performed best on PaviaU and Salinas (appearing at the top of Table~\ref{tab:baseline_methods}) while still ensuring coverage of the three prominent families: SNMF-sparsity and clustering, BSNETS-Reconstruction with Deep Learning, and Genetics-Supervised extensive searching with genetic optimization.

\subsection{Experimental Setting}
\noindent This subsection outlines the experimental settings used to evaluate our proposed models across the four datasets and two distinct tasks. The experimental procedures were tailored to accommodate the different characteristics of remote sensing and autonomous driving datasets.

\subsubsection{Experimental settings for the remote sensing task}
\label{sec:remote_sensing_experimental_settings}
\noindent For the remote sensing datasets (PaviaU, Salinas, and Chikusei), we employed a 10-fold cross-validation scheme, randomly splitting the pixels into training and validation folds. Hyperparameter tuning was first performed on the PaviaU dataset, and the derived settings were then applied to the Salinas and Chikusei datasets.

For the downstream classification task, we implemented a 3D Convolutional Neural Network (CNN) model, following the architecture outlined by Hamida et al.~\cite{8344565} and patch based pixel representation \cite{chen2016deep, he2017multi, DBLP:journals/corr/abs-1802-10478, DBLP:journals/corr/LeeK16, paoletti2023aatt, giri2024enhanced, zhang2024tree}. The network consists of a 1D convolutional layer followed by three 3D convolutional layers and a fully connected layer that outputs class probabilities. A batch size of 256 was used, and cross-entropy loss was employed. Our implementation was based on the publicly available code from Hamida et al.~\cite{Hyperspectral-Classification}.

For evaluation, we used overall accuracy as our primary metric. We also measured average accuracy and Kappa score, as they are commonly used to evaluate this task \cite{TAttMSRecNet, bs_nets}. 

\subsubsection{Experimental settings for the autonomous driving task}
\label{sec:autonomous driving_experimental_settings}
\noindent For the HSI-Drive v2 dataset, which involves image-level tasks, we split the 756 images into training, validation, and test sets using a 7:1:2 ratio. Hyperparameter tuning was performed on the validation set, and the final results were reported on the test set. We followed the experimental setup proposed by the original authors of the dataset, conducting one experiment for a 5-class scene understanding Advanced Driver Assistance Systems (ADAS) task.

For the downstream task, we implemented a UNet-based architecture that outputs semantic segmentation masks, following recent work~\cite{long2015fullyconvolutionalnetworkssemantic,  GUTIERREZZABALLA2023102878}. The encoder consists of 4 convolutional blocks, each applying a 0.5 downsampling factor to compress the input data. Each block utilizes a kernel size of 3 for the convolutional operations. The decoder mirrors this process by progressively upsampling the data to restore the original dimensions. As the loss function, we used Weighted Cross Entropy.

For the HSI-Drive v2 dataset, where each sample corresponds to an image, we evaluate performance using multiple metrics beyond Overall and Average Accuracy. Specifically, we report the Average Intersection over Union (IoU), Precision, and Recall for each label class, aligning with evaluation protocols used in similar settings \cite{GUTIERREZZABALLA2023102878}.

\subsubsection{EHBS hyperparameter settings}
\noindent For the band selection stage in the EHBS method, we set the batch size to 256. 
As proposed in \cite{pmlr-v119-yamada20a} \(\sigma\) and the initial values of the gating parameters \(\mu\) were set to 0.5.

For remote sensing data sets, we used the hyperparameters that worked best for the PaviaU data set. For the HSI-Drive v2 dataset, we used a validation set to determine good hyperparameters and reported the results on the test set; for it, we used a batch size of 16 images.
\subsubsection{CHBS hyperparameter settings}
\label{sec:CHBS_Hyperparameter_settings}
\noindent For CHBS (see~\cref{sec:concrete}), we initialize the model parameters $(L, \tau, \alpha, \beta)$. The temperature parameters ($\tau, \alpha$) and the noise parameters ($\beta$) were treated as hyperparameters, and their initialization values were tuned like other network hyperparameters (see \cref{sec:remote_sensing_experimental_settings,sec:autonomous driving_experimental_settings}). We tested different initial temperatures of $\tau \in [0,10]$ and $\alpha \in [0.99,0.99999]$. The values chosen and used for the final evaluation of the remote sensing datasets were $\tau=1.5$, $\alpha=0.99998$, $\beta=0.15$, and for the HSI-Drive v2 dataset, we used $\tau=8.5$, $\alpha=0.9999$, $\beta=0.15$.

The initialization of the logits matrix was performed using our novel initialization scheme (see \cref{sec:logits_initialization}) and compared to a naive initialization consisting of Xavier initializing the complete matrix uniformly in which each row of $L$ has values drawn from a normal distribution with a mean of $0$. We discuss this further in \cref{sec:analysis_and_discussion}.

\subsubsection{Compute and reproducibility}
\noindent All experiments were conducted on an NVIDIA GeForce GTX 1080 Ti, using Python 3.8 and PyTorch version 2.2.2~\cite{paszke2017automatic}. The HSI-Drive v2 dataset required approximately 400\,MB of GPU memory, while the remote sensing datasets required about 1\,GB. Each epoch took between 1 and 5 minutes, depending on the dataset. The source code and detailed implementation is publicly available on GitHub\footnote{\githuburl}.

\subsubsection{Experimental design and evaluation}

\noindent As most of the baseline methods are not embedded, we first ran the various band selection models as a first stage. Then, we used the selected bands from each method to train the downstream task model using the baseline downstream task model (CNN or UNet). This ensured a proper a fair and consistent comparison of the various band selection methods. 

For evaluation, we compared the accuracy of the various methods for various target numbers of bands. 
In addition to standard metrics such as overall accuracy, average accuracy, and Kappa score, we introduce a novel evaluation metric: the area under the bands-performance curve (AUC). Inspired by the AUC metric in binary classification (commonly used with ROC curves), this metric provides a single scalar summary of a band selection model's performance across various numbers of selected bands, enabling a holistic comparison of different methods.

\section{Results}
\subsection{Remote Sensing Results}
\begin{figure}[!htbp]
    \begin{center}
        \includegraphics[width=0.99\linewidth]{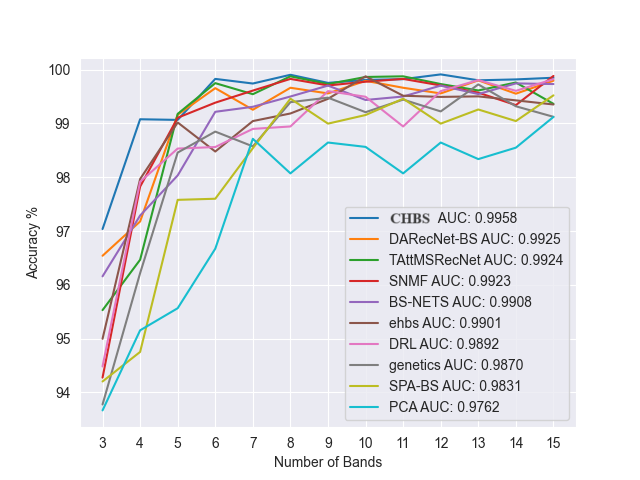}
    \caption{Overall accuracy and AUC results comparing our Gumbel-based method to other band selection methods for different numbers of selected bands over the PaviaU dataset}
    \label{fig:pavia_oa_a}
    \end{center}
\end{figure}

\begin{figure}[!htbp]
    \begin{center}
        \includegraphics[width=0.99\linewidth]{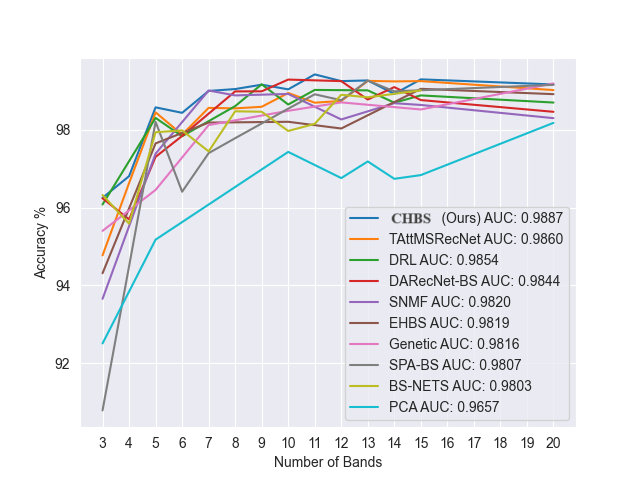}
    \caption{Overall accuracy and AUC results comparing our Gumbel-based method to other band selection methods for different numbers of selected bands over the Salinas dataset.}
    \label{fig:salina_oa_b}
    \end{center}
\end{figure}    
\noindent The overall accuracy results comparing our method to various baseline methods for different numbers of selected bands on the PaviaU and Salinas datasets are shown in Figures~\ref{fig:pavia_oa_a},\ref{fig:salina_oa_b}. Our proposed Gumbel-based method consistently outperforms the other methods for nearly all target numbers of bands in both datasets and achieves the highest AUC results, as also shown in the figures.

Table~\ref{tab:pavia_3_and_8_bands} presents the overall accuracy, average accuracy, and Kappa results for band selection on the PaviaU dataset with 3 and 8 selected bands. We performed 10-fold cross-validation for all remote sensing datasets (see \cref{sec:remote_sensing_experimental_settings}), and for each metric the standard deviation (std) across folds is reported after the $\pm$ symbol.  As highlighted in the table, our CHBS method outperforms all other methods in all three metrics when eight bands are selected. It also outperforms all other methods for Kappa and overall accuracy when 3 bands are selected; for average accuracy with 3 bands, our method is second best to BS-Net-Conv with very similar results.  

\begin{table*}[!htb]
\centering
\resizebox{\textwidth}{!}{ 
\begin{tabular}{|l|ccc|ccc|}
\hline
\multirow{2}{*}{Method/Metric} & \multicolumn{3}{c|}{3 bands} & \multicolumn{3}{c|}{8 bands} \\
 & Kappa	&	OA	&	AA & Kappa	&	OA	& AA\\ 
\hline			
CHBS	&	\textbf{0.97	 $\pm$	0.01}	&	\textbf{97.41	 $\pm$	0.46}	&	97.35	 $\pm$	0.63	&	\textbf{0.999	 $\pm$	0.001}	&	\textbf{99.90	 $\pm$	0.05}	&	\textbf{99.92	 $\pm$	0.58}\\
EHBS	&	0.93	 $\pm$	0.02	&	94.28	 $\pm$	1.65	&	94.49	 $\pm$	1.10	&	0.989	 $\pm$	0.006	&	99.19	 $\pm$	0.46	&	99.43	 $\pm$	0.42 \\
genetic	&	0.93	 $\pm$	0.01	&	94.59	 $\pm$	0.64	&	94.86	 $\pm$	0.99 &	0.992	 $\pm$	0.003	&	99.39	 $\pm$	0.26	&	99.54	 $\pm$	0.15		\\
TAttMSRecNet,	&	0.95	 $\pm$	0.01	&	96.50	 $\pm$	0.47	&	97.10	 $\pm$	0.63	&	0.991	 $\pm$	0.008	&	99.31	 $\pm$	0.60	&	99.33	 $\pm$	0.62\\
DARecNet-BS	&	0.95	 $\pm$	0.01	&	96.19	 $\pm$	0.98	&	97.15	 $\pm$	0.54
&	0.992	 $\pm$	0.009	&	99.38	 $\pm$	0.70	&	99.31	 $\pm$	0.70\\
BS-Net-Conv	&	0.95	 $\pm$	0.01	&	96.39	 $\pm$	0.60	&\textbf{97.45	 $\pm$	0.48}	&	0.994 $\pm$	0.010	&	99.57	 $\pm$	0.25	&	99.72	 $\pm$	0.12\\
DRL	&	0.92	 $\pm$	0.01	&	94.06	 $\pm$	0.91	&	93.86	 $\pm$	0.83	
&	0.991	 $\pm$	0.005	&	99.33	 $\pm$	0.35	&	99.35	 $\pm$	0.36\\
PCA	&	0.93	 $\pm$	0.01	&	94.74	 $\pm$	0.39	&	95.30	 $\pm$	0.67
&	0.973	 $\pm$	0.012	&	97.97	 $\pm$	0.91	&	98.01	 $\pm$	1.01\\
SpaBS	&	0.92	 $\pm$	0.01	&	93.68	 $\pm$	0.46	&	94.29	 $\pm$	0.54
&	0.994	 $\pm$	0.002	&	99.55	 $\pm$	0.23	&	99.58	 $\pm$	0.20\\
SNMF	&	0.91	 $\pm$	0.02	&	93.42	 $\pm$	1.39	&	93.03	 $\pm$	1.68
&	0.992	 $\pm$	0.012	&	99.39	 $\pm$	0.92	&	99.44	 $\pm$	0.81\\
\hline
\end{tabular}
}
\caption{Overall Accuracy (OA), Average Accuracy (AA), and Kappa results for the PaviaU dataset using 3 and 8 selected bands. \\Results are averaged over 10-fold cross-validation with standard deviations reported.}
\label{tab:pavia_3_and_8_bands}
\end{table*}

\begin{figure*}[!htbp]
    \begin{center}
    \begin{subfigure}{0.48\textwidth}
        \includegraphics[width=\linewidth]{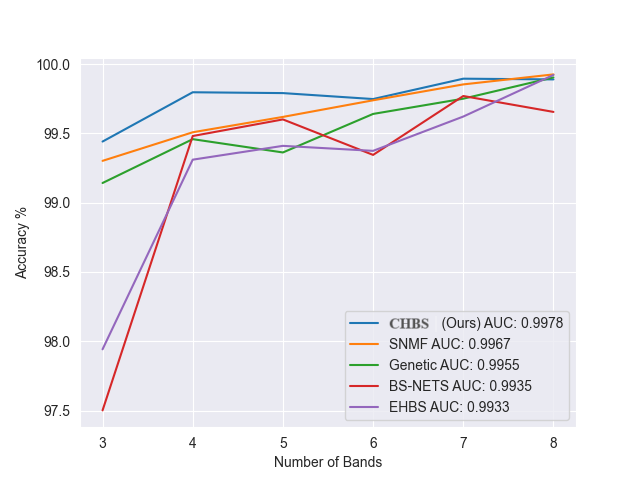}
        \caption{Overall accuracy and AUC results for Chikusei.}
        \label{fig:chikusei_oa}
        \vspace{1em}
    \end{subfigure}
    \hfill 
    \begin{subfigure}{0.48\textwidth}
        \includegraphics[width=\linewidth]{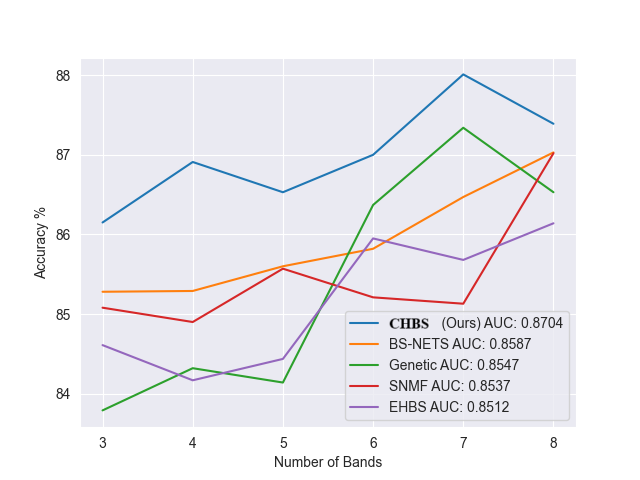}
        \caption{Mean IOU for HSI-Drive-v2 with 5 classes}
        \label{fig:drive5-mean-iou}
        \vspace{1em}
    \end{subfigure}    
\caption{Results for (a) Chikusei and (b) HSI-Drive-v2 datasets}
\label{fig:chikusei_and_drive_results}
\end{center}
\end{figure*}

The overall accuracy and AUC results for the Chikusei dataset are shown in Figure~\ref{fig:chikusei_oa}. We have tested our method and compared it to the top-performing band selection methods on the PaviaU and Salinas datasets. Results show that our proposed concrete method outperforms all other methods in overall accuracy on a low number of selected bands as well as in the overall AUC.

\subsection{Autonomous Driving Results}
\noindent Mean IOU results for the HSI-Drive v2 dataset with a 5-class classification are shown in Figure~\ref{fig:drive5-mean-iou}. Our method consistently outperforms all other baseline methods across all target numbers of bands, achieving higher segmentation quality and overall AUC. Table~\ref{tab:drive_3_and_8_bands_results} shows detailed results for the HSI-Drive v2 dataset for 3 bands and 8 bands selected, comparing our proposed concrete method to others over 12 evaluation metrics. As can be seen, our proposed concrete method outperforms all other methods in all evaluation metrics when 3 bands are selected and in all but 2 metrics when 8 bands are selected (for these two cases, our method is ranked as a close second with a low margin).


\begin{table*}[!htbp]
\centering
\resizebox{\textwidth}{!}{ 
\begin{tabular}{|l|ccccc|ccccc|}
\hline
\multirow{2}{*}{Method/Metric} & \multicolumn{5}{c|}{3 bands} & \multicolumn{5}{c|}{8 bands} \\
 &SNMF &BSNETS &Genetic & EHBS& CHBS & SNMF &BSNETS &Genetic & EHBS& CHBS\\ 
\hline			

Mean IoU & 85.08 & 85.28 & 83.79  &84.61&\textbf{86.15} & 87.02 & 87.03&  86.53& 86.14&\textbf{87.39}\\
Overall IoU & 92.99 & 93.24 & 92.41 &92.65& \textbf{93.59} & 94.06& 94.08&93.83&93.54&\textbf{94.24} \\ 
Weighted IoU & 93.53 & 93.76 & 93.01& 93.28 & \textbf{94.04} & 94.48&94.50& 94.33&94.04&\textbf{94.66}\\
Mean Precision & 90.18  & 90.34 & 88.72 & 89.79&\textbf{91.15} & 90.91& 91.40 &91.01&90.92&\textbf{91.59}\\
Mean Recall &92.63  & 92.88 & 93.00&92.28 &\textbf{93.35} & \textbf{94.55}&93.91&93.56&93.05&94.18\\
Overall Accuracy &96.37  & 96.50 & 96.06 &96.19 &\textbf{96.69} & 96.94&96.95&96.82& 96.66&\textbf{97.03}\\
Mean Accuracy &  92.63& 92.88 &  93.00&92.28 & \textbf{93.35} & \textbf{94.55}&93.91&93.56&93.04&94.18\\
\hline
\end{tabular}
}
\caption{Detailed results for the HSI-Drive v2 dataset with 3 and bands 8 selected}
\label{tab:drive_3_and_8_bands_results}
\end{table*}

\subsection{Analysis and Discussion}
\label{sec:analysis_and_discussion}
\noindent Our experimental results indicate that the proposed concrete selector method consistently outperforms existing state-of-the-art approaches, particularly when only a small number of spectral bands are selected. In contrast, while the EHBS method achieves performance comparable to other SOTA techniques, its overall results fall short of those obtained with the concrete-based CHBS method. These findings suggest that dynamically learning band importance via the Gumbel-Softmax reparameterization offers superior task-specific feature selection, paving the way for more efficient hyperspectral analysis in real-world applications.
 
We attribute the superior performance of the concrete method to its ability to learn a specific set of bands tailored to the given task, 
effectively discarding redundant or less relevant spectral regions (e.g., extreme bands that are hard to acquire) that do not contribute significantly to the task performance. 
The CHBS model requires a number of learnable parameters equal to the product of the total number of bands and the number of target bands. Thus, reducing redundancy in this layer is key to efficient learning. As a future direction, we plan to explore methods for further reducing the parameter count in the band selection layer to improve scalability for large-scale applications.

\subsubsection{Selected Bands Analysis}
\begin{figure}[!htbp]
    \begin{center}
        \includegraphics[width=0.99\linewidth]{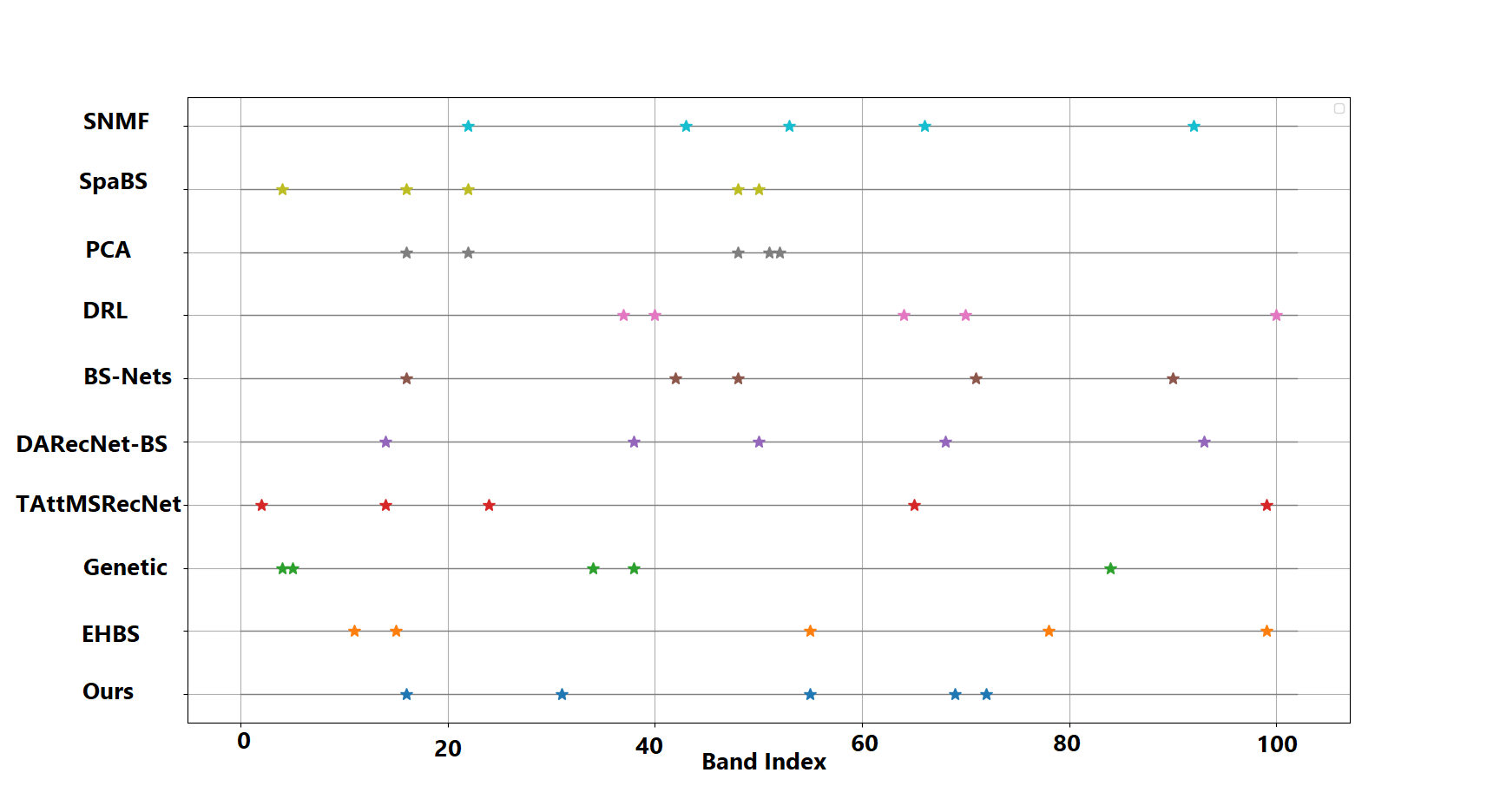}
        \caption{Illustration of the 5 selected bands for PaviaU by different models}
        \label{fig:selected_bands_pavia}
    \end{center}
\end{figure}

\noindent Figure~\ref{fig:selected_bands_pavia} compares the bands selected by our proposed concrete method to those selected by other methods on the PaviaU dataset. We observe that reconstruction-based methods tend to select bands that cover the extremes of the spectral range as they are optimized for overall data reconstruction. In contrast, our method—by focusing on the downstream task performance—selects a more compact and task-specific set of bands that leads to improved classification accuracy without relying on these extreme and often noisier spectral regions.

\begin{figure}[!htbp]
    \begin{center}
        \includegraphics[width=0.99\linewidth]{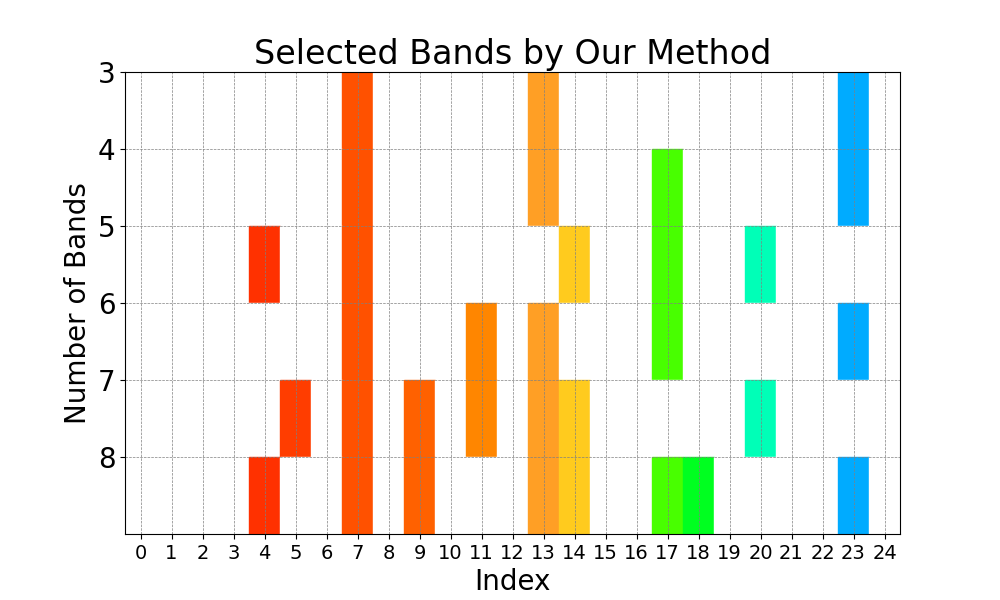}
        \caption{Illustration of the bands selected by our method for HSI-Drive-V2 with different numbers of target bands.}
        \label{fig:our_hsi_drive_5}
    \end{center}
\end{figure}

Figure~\ref{fig:our_hsi_drive_5} shows the selected bands for the HSI-Drive v2 dataset using our method for different target numbers of bands. The selection is consistent, with the top 3 bands (top of figure) remaining largely unchanged as the number of target bands increases to 8 (bottom row of figure), which reinforces the stability of our approach.

\subsubsection{Impact of Segmented Xavier Initialization on CHBS} 

\begin{figure}[!htbp]
    \begin{center}
    \includegraphics[width=0.98\linewidth]{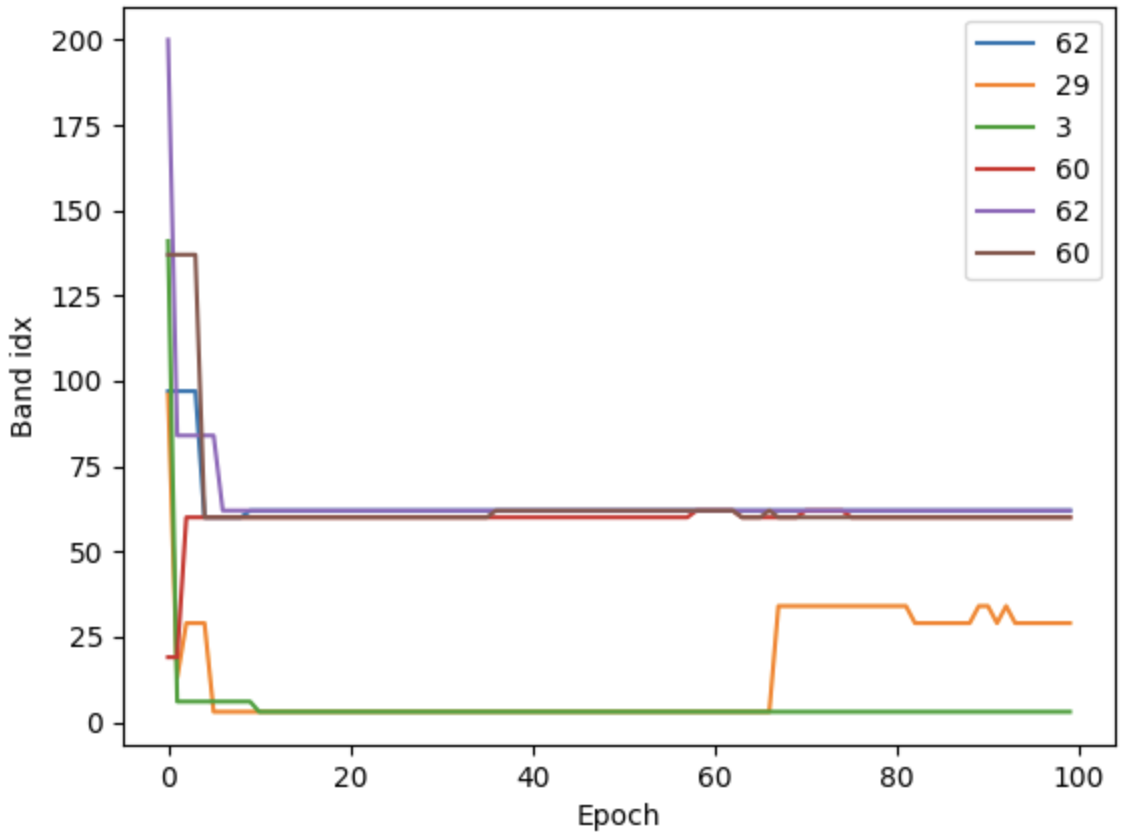}
    \caption{CHBS with random initialization: band selection progression on the Salinas dataset with 6 bands.}
    \label{fig:collapsed_encoder}
    \end{center}
\end{figure}

\begin{figure}[!htbp]
    \begin{center}
        \includegraphics[width=0.98\linewidth]{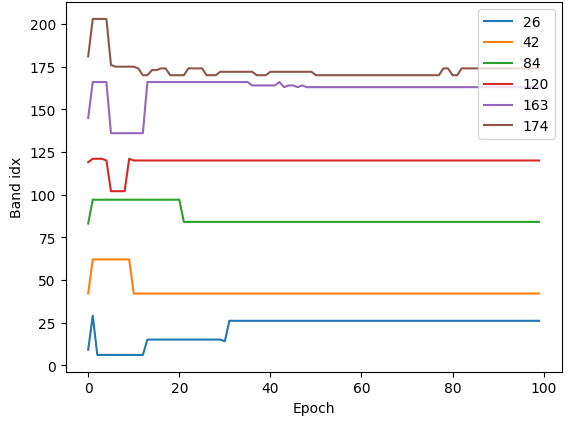}
        \caption{CHBS with Segmented Xavier Initialization: band selection progression on the Salinas dataset with 6 bands.}
        \label{fig:our_encoder}
    \end{center}
\end{figure}

\noindent In \cref{sec:logits_initialization}, we introduced a novel initialization scheme, Segmented Xavier Initialization, designed to mitigate band duplication and improve the diversity of selected bands in the CHBS encoder. This subsection presents an analysis demonstrating the effectiveness of our proposed initialization method compared to standard approaches.
During initial experiments, we observed that CHBS with standard Xavier uniform initialization often led to the repeated selection of certain bands, reducing the overall diversity of the selected set. Figures~\ref{fig:collapsed_encoder},\ref{fig:our_encoder} illustrates the progression of band selection on the Salinas dataset when targeting 6 bands. In Figure~\ref{fig:collapsed_encoder}, we show the selection behavior with naive Xavier uniform initialization of the logits matrix. As training progresses, the selection collapses, converging to only 3 distinct bands instead of fully utilizing the available spectral range.
To address this issue, we applied the Segmented Xavier Initialization, which strategically biases the initialization towards distinct spectral regions, as described in \cref{sec:logits_initialization}. Figure~\ref{fig:our_encoder} demonstrates that with our improved initialization, the encoder successfully selects 6 diverse bands across the spectral range, preventing collapse and ensuring a well-distributed band selection. 

\begin{figure}[!htbp]
    \begin{center}
    \includegraphics[width=0.5\linewidth]{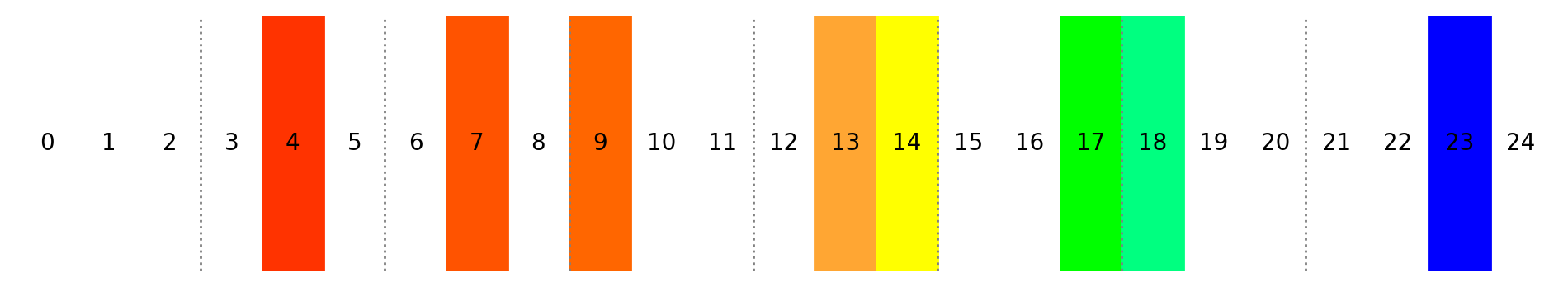}
    \caption{Selected bands and corresponding initialization areas for the HSI-Drive v2 dataset with 8 bands.}
    \label{fig:initialization-areas}
    \end{center}
\end{figure}

Our novel initialization scheme significantly enhances overall accuracy while avoiding band duplication. This is achieved by initializing each selector to focus on a different region of the spectrum, promoting a diverse selection of bands and ultimately improving model performance. Notably, the learning process is not constrained to selecting a single band per segment. For instance, Figure~\ref{fig:initialization-areas} shows the selected bands and corresponding initialization areas for the HSI-Drive v2 dataset when selecting 8 bands from 25. As observed, the method did not select any bands from the first segment (bands 0-2) but chose two bands from another region (bands 12-14), demonstrating its flexibility.

We also experimented with alternative initialization strategies, including seeding band indices selected by other methods, such as SNMF SpaBs and BSNets with higher values. While this approach improved selection stability, it did not perform as well as our segment-based initialization. Additionally, our method does not rely on running other selection techniques, making it the preferred choice for our setting.


\section{Conclusion}
\noindent We introduced two embedded methods for hyperspectral band selection: \textbf{EHBS}, using structured $\ell_0$-regularization, and \textbf{CHBS}, based on a Concrete Selector Layer with Gumbel-Softmax reparameterization. Both integrate band selection directly into model training, enabling efficient, task-specific feature selection without separate preprocessing.

Experiments on three remote sensing benchmarks and an autonomous driving dataset show that both methods achieve strong performance, with CHBS excelling when selecting few bands—a key advantage for real-time, resource-limited applications.

A limitation of our approach is its sensitivity to hyperparameters such as learning rate and regularization weight, which varied across datasets and tasks. This highlights the need for improved robustness and easier deployment.
Our findings demonstrate the practical potential of embedded band selection in deep learning pipelines, paving the way for more efficient hyperspectral imaging in applications such as precision agriculture, environmental monitoring, and autonomous navigation.





\begin{ack}
\noindent This work was partially supported by the Chief Scientist of the Israeli Ministry of Agriculture grant number 12-03-0010. 
\end{ack}

\FloatBarrier


\bibliography{mybibfile}

\end{document}